\documentclass[runningheads]{llncs}
\usepackage[T1]{fontenc}
\usepackage{graphicx}
\usepackage{subfigure}

\usepackage{xcolor}
\usepackage{amsmath}
\usepackage{amssymb}
\usepackage{hyperref}

\begin{document}
	
	\title{A lightweight method for detecting dynamic target occlusions by the robot body}
	\titlerunning{A lightweight method for detecting occlusions}
	
	\author{Savvas Sampaziotis \and
		Sotiris Antonakoudis \and
		Marios Kiatos \and
		Fotios Dimeas \and
		Zoe Dougleri
	}
	
	\authorrunning{S. Sampaziotis et al.}
	\institute{Aristotle University of Thessaloniki, Greece \\
		\email{\{sampazio,antonakos,dimeasf,doulgeri\}@ece.auth.gr}}
	
	\maketitle   
	
	\begin{abstract}
		
		Robot vision is greatly affected by occlusions, which poses challenges to autonomous systems. The robot itself may hide targets of interest from the camera, while it moves within the field of view, leading to failures in task execution. 
		For example, if a target of interest is partially occluded by the robot, detecting and grasping it correctly, becomes very challenging. To solve this problem, we propose a computationally lightweight method to determine the areas that the robot occludes. For this purpose, we use the Unified Robot Description Format (URDF) to generate a virtual depth image of the 3D robot model. Using the virtual depth image, we can effectively determine the partially occluded areas to improve the robustness of the information given by the perception system. Due to the real-time capabilities of the method, it can successfully detect occlusions of moving targets by the moving robot. We validate the effectiveness of the method in an experimental setup using a 6-DoF robot arm and an RGB-D camera by detecting and handling occlusions for two tasks: Pose estimation of a moving object for pickup and human tracking for robot handover. The code is available in \url{https://github.com/auth-arl/virtual\_depth\_image}.
		
	\end{abstract}
	
	\section{Introduction}
	
	Robotic systems that need to operate in unstructured environments and to present versatility in their operation, such as for object pickup, grasping in cluttered environments, and avoiding obstacles, require machine vision for enhanced perception capabilities. This is achieved with cameras that cover the robot's workspace and provide the necessary information that is critical to the robot's operation. One of the biggest problems, however, is occlusions. These are blind spots of targets of interest in the camera view. A target may be an object that the robot needs to manipulate (e.g., for pick \& place) or a human in the workspace that the robot needs to interact with. Such blind spots can have a negative impact on the performance of the perception methods. An error in perception can have a serious impact on the robot's performance, preventing it from completing its task successfully and risking damage from a possible misbehavior.
	
	Perception systems use cameras in various configurations to reduce occlusions, depending on the application. For example, object detection systems for pickup most commonly use overhead cameras \cite{wuthrich2013probabilistic}. Human tracking systems use cameras mounted at an angle rather than overhead to keep the human body in front view and cover a larger field of view towards the collaborative workspace \cite{Magrini2014}. One way to reduce occlusions is to use multiple cameras with different viewing angles \cite{melchiorre2019collison}. However, depth-sensing cameras with structured light or lidar technologies can be quite expensive and interfere with each other. In addition, image processing of additional cameras multiplies computational costs. A common method to prevent occlusion is camera-in-hand configuration. However, this is not always practical, for example, when the entire working area must be covered or the measurement distance is less than the minimum allowable distance of the depth sensor. Moreover, motion blur is introduced by the moving camera. 
	
	Occlusion detection has been studied in the literature to improve the performance of perception systems for object tracking.
	In \cite{yoshioka2021through}, a simple approach is proposed to detect occluded areas in the sensing data (e.g., point cloud) using a predefined height threshold. When the robot arm is above the target objects that are laid on a table, any point above the threshold is classified as occluded. This approach has significant limitations in the operation of the robot and in the flexibility of the setup, since the classification is based on a division surface. The authors in \cite{luo2021calibration} address occlusions of target objects from the robot by training a deep neural network with a dataset containing images of occluded objects by the robot arm. However, this method requires a large amount of data to be trained and the inference time can be considerably high.
	In \cite{wuthrich2013probabilistic}, an object tracking method handles occlusion of a tracked object by comparing the depth measurement from the camera with the known 3D model of the object. A drawback of this approach is its dependence on an object tracking algorithm. Accuracy errors in object tracking affect the accuracy of the occlusion detection method. In addition, it cannot be used with methods where the 3D model is unknown or in other applications such as human tracking.
	
	Occlusion detection specifically for human tracking is an active research topic and has the potential to improve the safety aspect of human-robot collaboration in shared workspaces. The authors in \cite{isobe2018occlusion} propose an occlusion handling method for a mobile robot that follows humans using color and disparity information from a stereo camera. However, it depends heavily on the discrete depth information capability to detect occlusions, which may affect the accuracy in close proximity to the target from the occlusion.
	Shu et al. \cite{shu2012part} proposed a method for detecting occlusions of human body parts in skeleton tracking. Although this method can obtain regional features, it scans the scene where the features are extracted, which requires high computational cost. 
	To detect occlusions between overlapping individuals in an image, Cielniak et al. \cite{cielniak2007improved} used color, depth, and thermal images to train a classification algorithm. 
	
	Information about the structure of the robot, such as the URDF model, has been used in the literature to improve perception. URDF can be used to model robot arms using a hierarchical structure with XML trees and is well supported in Robotics Operating System (ROS). It can contain information about the kinematics, link inertia values, visual representation, and even the collision model of a robot, which is usually a slightly enlarged simplified body of each link. URDF filtering \cite{urdfFilter} was used in \cite{Magrini2014} along with an RGB-D camera to avoid ambiguities in tracking the human skeleton when the hands get very close to the robot. However, URDF filtering does not distinguish whether the targets are actually occluded or whether they are between the line of sight from the camera to the robot, because the area of the robot model is subtracted from the depth image, causing loss of useful depth information.
	To address this, the authors in \cite{fetzner20143d} fused the depth images from two separate sensors. Besides the cost of the additional sensor, the robot model had to be rendered twice from the perspective of each camera and then fused, increasing the computational cost.
	
	This work proposes a computationally lightweight method for determining the areas that the robot occludes from the camera. Instead of subtracting the robot model from the depth image, we generate a virtual depth image from the URDF model with minimal computational resources, and then fuse this virtual depth image with the depth image captured by the camera sensor to determine the area that the robot actually occludes. Our method runs in real time with a single RGB-D camera and is independent of the application or perception system. In the experiments, we demonstrate the proposed occlusion detection and how occlusions can be handled in two different use cases.
	
	\section{Methodology}
	
	Consider a robot receiving input from a perception system that detects and tracks targets of interest related to its task. The perception system uses an RGB-D camera with the robot being either fully or partially in its field of view. During task execution the robot may partially or fully occlude the targets, which may affect the target detection accuracy of the perception system.
	
	\begin{figure}[!b]
		\centering
		\includegraphics[width=.75\columnwidth]{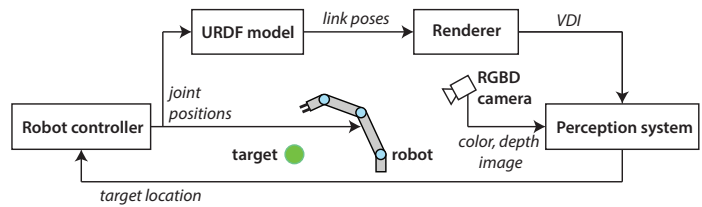}
		\vspace*{-3mm}
		\caption{Block diagram of the architecture that utilizes the virtual depth image (VDI).}
		\label{fig:block}
	\end{figure}
	
	Having the URDF model and the transformation of the camera's coordinate frame from the robot base, we can use the joint position measurements of the robot to render the virtual depth image (VDI). The VDI is a noise-free depth image of the robot from the camera's point of view, from which a perception system can infer whether the target is partially or fully occluded, improving the robustness of the perception system. The block diagram in Fig. \ref{fig:block} shows a general architecture that uses the VDI and indicates that the method is independent of the perception system.
	
	\subsection{Rendering the Virtual Depth Image}
	
	The first step of the method is to render depth images of the robot from the camera's point of view. We use a virtual camera with a pinhole camera model, where the intrinsic and extrinsic parameters correspond to the known parameters of the actual camera. 
	The image distortion of the actual camera is compensated to match the pinhole camera model.
	To render the image, the 3D model is taken from the URDF and the joint positions of the actual robot. 
	The 3D model need not be an exact representation of the robot, as fine details will have a negligible effect on the final image for occlusions. For this purpose, we use the collision model, which is a convex approximation of each robot link in Standard Triangle Language (STL). The geometry of each link is represented by a mesh of triangles defined by a set of vertices. The use of the collision model is a conservative approach for detecting occlusions, because it increases the area marked as occluded in the VDI, but on the other hand it can compensate for small calibration errors of the camera.
	
\begin{figure}[!t]
		\centering
		\subfigure[]{\includegraphics[width=0.47\textwidth,trim={0 20px 0 20px},clip]{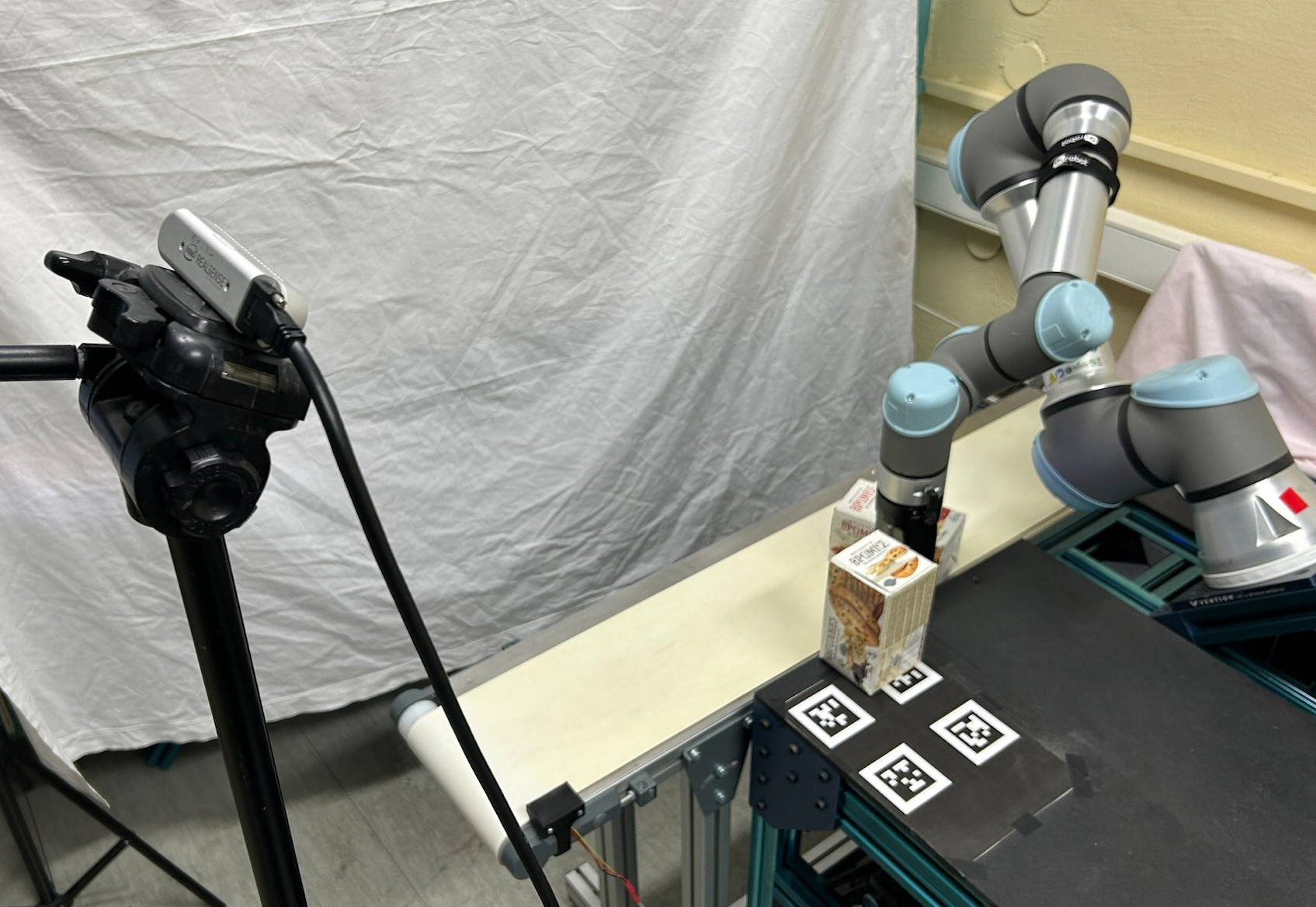}}
		\subfigure[]{\includegraphics[width=0.47\textwidth,trim={0 60px 0 60px},clip]{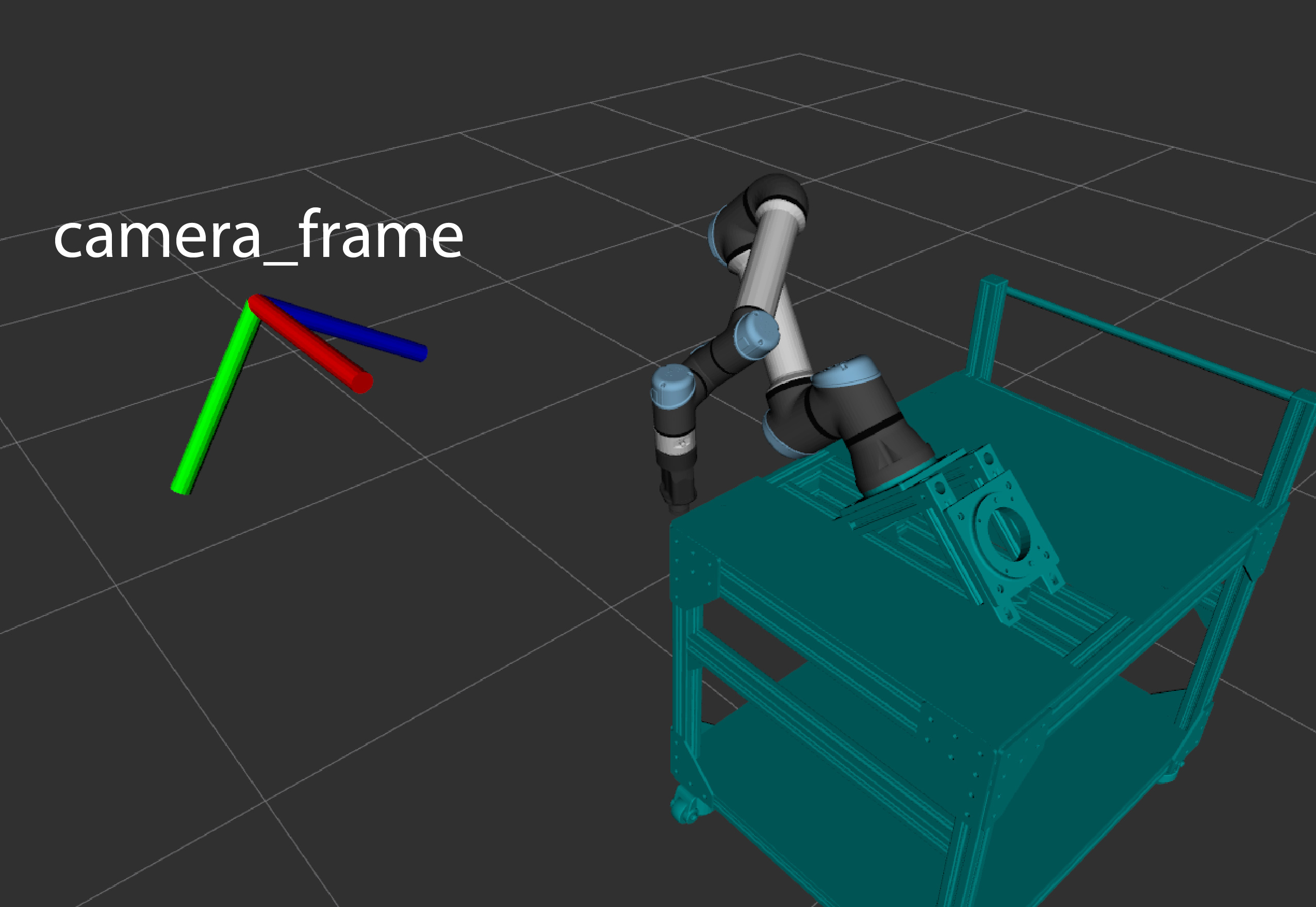} \vspace*{-5mm}}
		\subfigure[]{\includegraphics[width=0.31\textwidth,trim={0 40px 0 0px},clip]{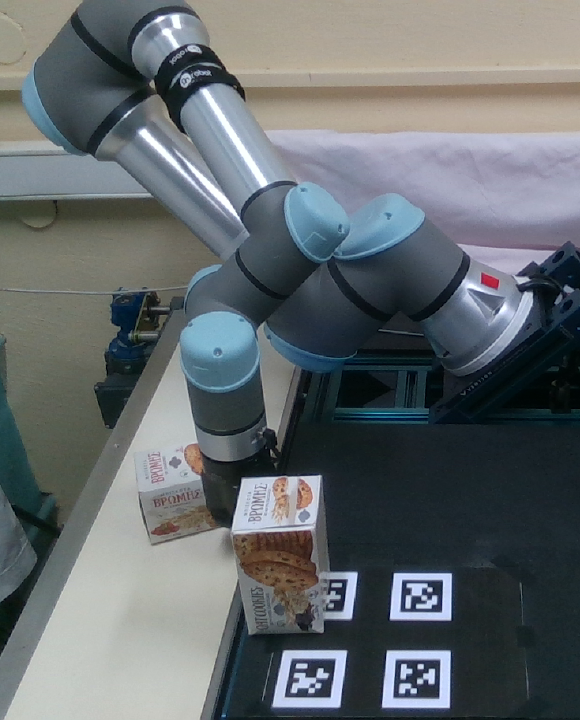}}
		\subfigure[]{\setlength{\fboxsep}{0pt}%
			\fbox{\includegraphics[width=0.31\textwidth,trim={0 40px 0 0},clip]{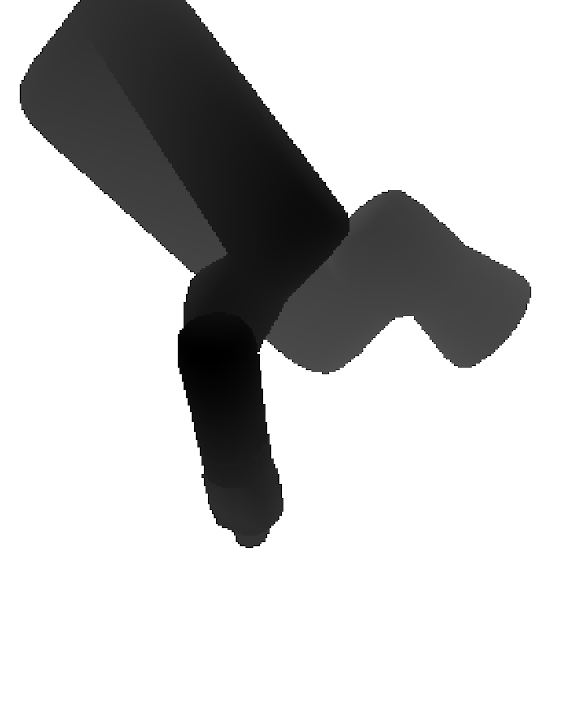}}}
		\subfigure[]{\includegraphics[width=0.31\textwidth,trim={0 40px 0 0},clip]{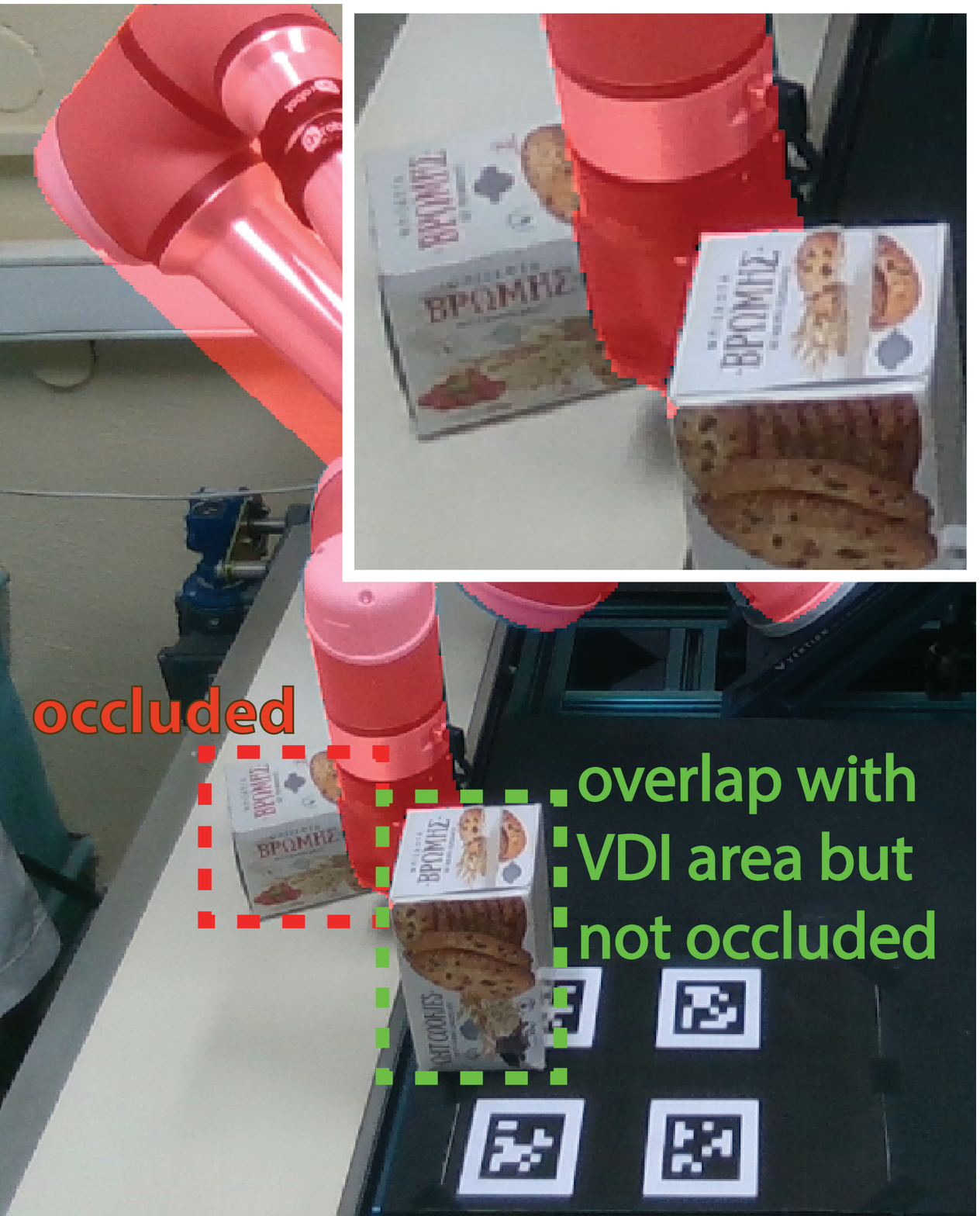}}
		\vspace*{-3mm}
		\caption{(a) Setup with a robot, an RGB-D camera and two objects (one in front and one behind the robot), (b) the digital visualization, (c) the color view from the camera, (d) the rendered VDI including the end-effector, (e) the augmented view highlighting the areas occluded by the robot.}
		\label{fig:overview}
	\end{figure}

	For the implementation of the renderer we used the OpenGL library. This library allows us to build a simple graphics pipeline that takes advantage of a computer's graphics processor and achieve a high frame rate while using minimal CPU resources. The frame rate of the virtual camera must match that of the real camera to produce a matching image pair. The renderer first obtains the joint positions of the robot, then transforms the pose of each robot link from the world frame to the camera frame and finally renders all 3D models to generate each image. The output is a noise-free virtual depth image of the robot and any other geometry in the URDF we choose to render, as shown in Fig. \ref{fig:overview}.
	
	The implemented graphics pipeline is described by the MVP (Model-View-Projection) homogeneous transformation matrix $\mathbf{g}_{MVP} \in \mathrm{R}^{4 \times 4}$, which is applied on all vertices of each $i^{th}$ robot link to recreate it from the camera view: 
	\begin{equation}
	\mathbf{g}_{MVP} = \mathbf{g}_P \mathbf{g}_V \mathbf{g}_M^i \ .
	\end{equation}
	The model matrix $\mathbf{g}_M^i$ is unique for each 3D object rendered in the scene. It is defined by each robot link's pose in respect to world frame and is derived from the robot's forward kinematic model and joint positions.
	The view matrix $\mathbf{g}_V$ is the homogeneous transformation from the world frame to the camera frame, which can be obtained from a camera registration procedure.    
	The camera projection matrix $\mathbf{g}_P$ transforms the vertices into the camera space. The matrix parameters are  defined by the actual RGB-D camera intrinsics:
	\begin{equation}
	\mathbf{g}_P = \begin{bmatrix}
	\frac{f_x}{c_x} & 0 & 0 & 0 \\
	0 & \frac{f_y}{c_y} & 0 & 0 \\
	0 & 0 & -\frac{f+n}{f-n} & -\frac{2 f n}{f-n} \\
	0 & 0 & -1 & 0 \\ 
	\end{bmatrix},
	\end{equation}
	where $f_x,f_y$ and $c_x, c_y$ are the focal lengths and principal point of the pinhole model of the actual depth camera. 
	The $f$ and $n$ parameters are called the "far" and "near" plane and refer to the maximum and minimum depth that will be rendered. Their values are chosen to match the actual depth camera range.

	\subsection{Occlusion detection and handling}
\begin{figure}
		\centering
		\includegraphics[width=0.7\columnwidth]{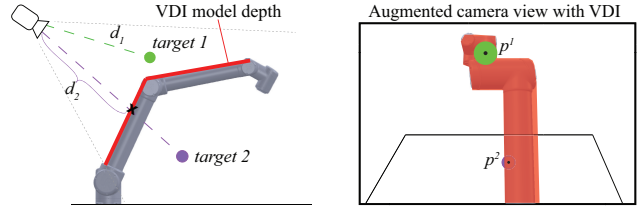}
		\vspace*{-5mm}
		\caption{Two targets that lie within the VDI area. Without the occlusion mask, de-projecting the partially occluded target 2, results in wrong distance $d_2$. Target 1 is correctly detected and measured despite being within the VDI area.}
		\label{fig:projections}
	\end{figure}

	Let $p_{ij}$ be a pixel at an area of interest on the color image of the RGB-D camera with coordinates $i,j$. By comparing the actual and virtual depth measurements $d_{ij}$ and $d^V_{ij}$ respectively, we can deduce whether this pixel is occluded by the robot and whether it can be safely de-projected to 3D space using the depth measurements, as shown in Fig. \ref{fig:projections}. 
	Specifically, if $d_{ij} < d^V_{ij} - \epsilon$ then the point of interest is between the camera and the robot and no occlusion occurs. $\epsilon$ is a small positive value that compensates for the depth measurement noise and calibration errors. On the contrary, if $d_{ij} \geq d^V_{ij} - \epsilon$ then the depth measurement is derived from the robot's body and the point of interest is occluded. 
	
	The way to handle an occlusion after it has been detected depends on the method of the perception system. Different methods for target detection and tracking have different robustness to the occlusion percentage of the target \cite{kuipers2020hard}. In pose estimation of a moving object, for example, when the occlusion percentage of the object is higher than the acceptable occlusion threshold, future poses of the object can be predicted based on the current non-occluded estimates of pose and velocity \cite{chandel2015occlusion}. 
	
	\section{Experimental evaluation}
	
	To evaluate the proposed method, we conducted two experiments using a UR5e 6-DOF robot arm and an Intel Realsense D415 RGB-D camera.
	In the first experiment, we demonstrate how we can handle occlusions that affect the pose estimation of a moving object for robot pickup. In the second one, we show a successful handover task in which the detected human is occluded by the robot during task execution. Our method is implemented with ROS2 on a Ryzen 3700x CPU and a GTX1060 GPU. The virtual depth image of the robot arm is rendered real time at 30 fps rate, equal to the camera frame rate. The average calculation time of each VDI frame is less than 10ms. The values of the parameters in the experiments are $n=0.5$m, $f=3$m, $\epsilon=0.01$m.

	\subsection{Pose estimation of a moving target object for pickup}
	
	Robot pickup of objects moving on a conveyor belt requires a perception system that estimates the pose of the target object. In this experiment, we use a perception system based on Mask-RCNN network \cite{he2017mask}, fine-tuned on synthetic images, to detect the mask of the target object and a subsequent feature matching method to estimate the 6-DOF pose of the target based on SIFT features \cite{tang2012textured}. This two-stage approach is very common for estimating the pose of an object \cite{zaidi2022survey}. 
	However, during task execution, the robot may partially occlude the moving object, leading to wrong estimates of its pose.
	
\begin{figure}[!h]
		\centering
		\subfigure[]{\includegraphics[width=0.2\textwidth]{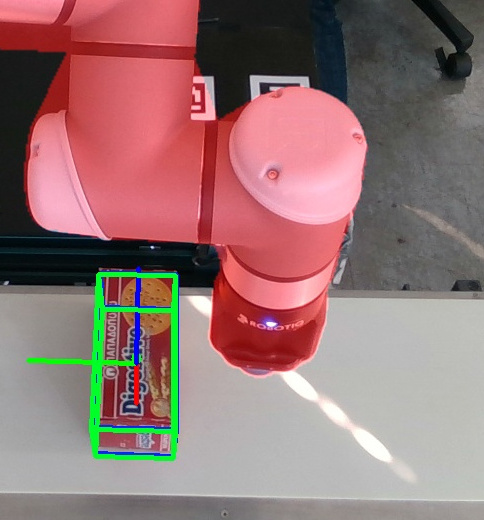}}%
		\subfigure[]{\includegraphics[width=0.2\textwidth]{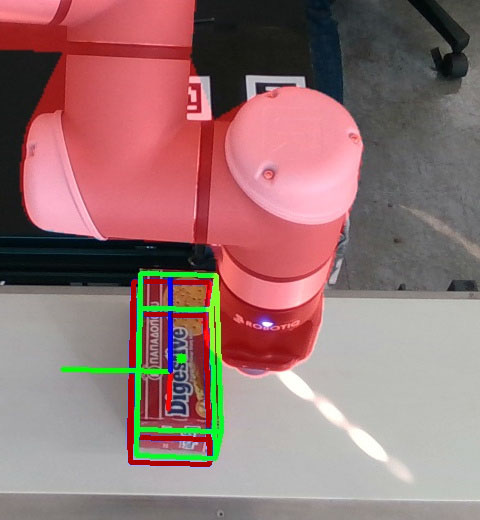}}%
		\subfigure[]{\includegraphics[width=0.2\textwidth]{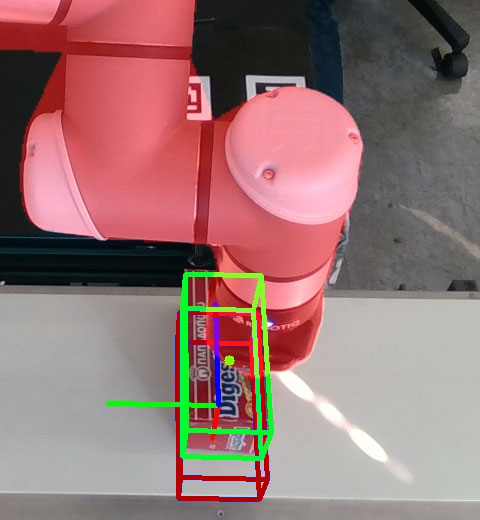}}%
		\subfigure[]{\includegraphics[width=0.2\textwidth]{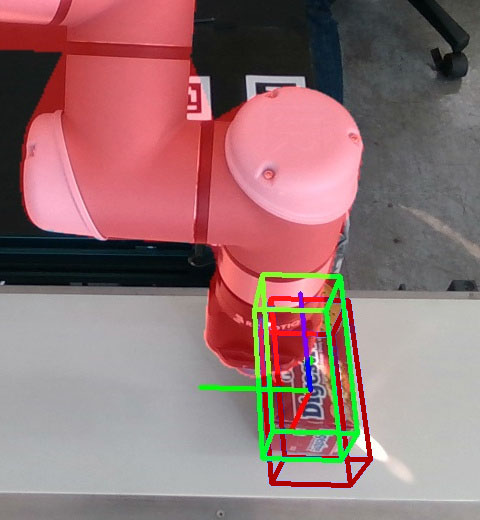}}%
		\subfigure[]{\includegraphics[width=0.2\textwidth]{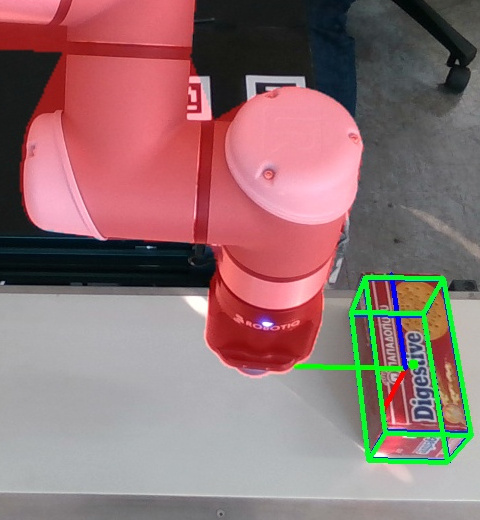}}
		\vspace*{-5mm}
		\caption{Pose estimation of a moving object on a conveyor. When the object is partially occluded by the robot, the object detection systems provides inaccurate estimates of the object's pose ({\color{red}red} bounding boxes). Utilizing the VDI, we can accurately track the object ({\color{green}green} bounding boxes) based on its position and velocity.}
		\label{fig:object_tracking}
	\end{figure}

	To deal with this problem, the proposed VDI is used with an object tracking method that predicts the pose of the object when the occlusion threshold is higher than 5\%, based on the last estimates of the pose and the velocity of the object before it got occluded. 
	When the percentage drops below that threshold, the tracker uses again the estimates from the pose detection method as described above. With this approach, the object can be accurately tracked despite the wrong pose estimates during partial occlusions, as shown in of Fig. \ref{fig:object_tracking}.

	\subsection{Human pose estimation for robot handover}
\begin{figure}
		\centering
		\subfigure[]{\includegraphics[width=\textwidth]{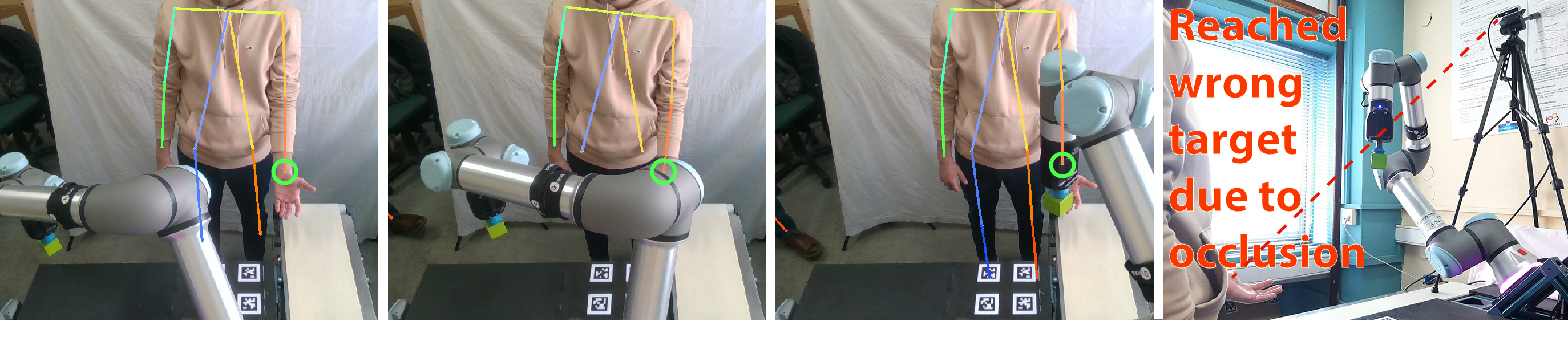} \vspace*{-5mm}}
		\subfigure[]{\includegraphics[width=\textwidth]{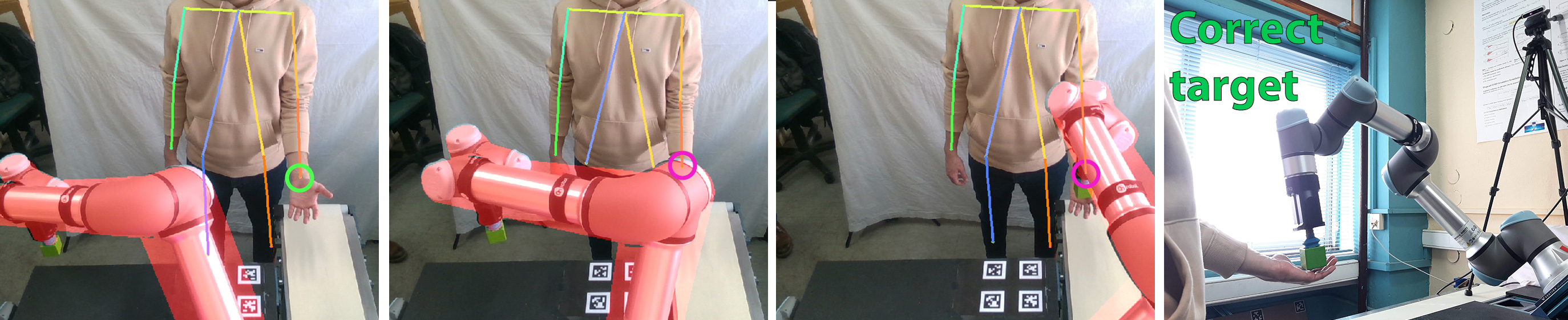}}
		\vspace*{-5mm}
		\caption{Handover experiment where the robot reaches the detected left hand of the user. (a) With simple de-projection the robot reaches the wrong target because of occlusion. (b) With the VDI occlusions are handled and the robot reaches the correct target. } 
		\label{fig:handover}
	\end{figure}

	The proposed method is also applied to a handover task. In particular, the robot should give an item to the left hand of a human as shown in Fig. \ref{fig:handover}. To achieve that a perception system detects the wrist of the human as the target location for the handover. The AlphaPose algorithm \cite{fang2017rmpe} is used to detect the human skeleton 2D keypoints on the RGB image. Using the depth image, we de-project the keypoint that corresponds to the left wrist to acquire their 3D position. During the handover, the target position is continuously updated in case the human moves their hand. However, during the motion of the robot, occlusions occur as the ones depicted in Fig. \ref{fig:handover}. For example, the robot's elbow occludes the human hand while it is approaching the human and the robot reaches the wrong target position at the end of the handover due incorrect de-projection, as shown in Fig. \ref{fig:handover}a.  This happens because the human pose estimation algorithm predicts correctly the keypoint that corresponds to the left wrist of the human, even though the human is partially occluded. However, the de-projection of that 2D keypoint provides a 3D position that lies on the robot's elbow instead of the human hand.
	Using the VDI we can detect that this point is occluded and then handle the occlusion by using the last non-occluded target position, as shown in Fig. \ref{fig:handover}b. In this case, the robot reaches the correct target for handover, right over the hand of the user. Compared to relying on noisy depth measurements to handle occlusions, the VDI provides a mechanism to make perception capabilities more robust for robotic integration.
	
	\section{Conclusions}
	In this work we presented a method to detect occluded areas in the view of a camera by rendering the URDF. The generated Virtual Depth Image is noise-free, is calculated in real-time and can be used by a perception system to detect if a target is occluded by the robot or not, even if it overlaps with the VDI area. We demonstrated the effectiveness of the method in two use-cases that utilized a perception system for object tracking and another for human-tracking. The results suggest that this can be an effective way to detect occlusions and treat them appropriately, regardless of the perception method that is used.
	
	\subsubsection{Acknowledgements} 
	Co‐financed by the European Regional Development Fund of the European Union and Greek national funds through the Operational Program Competitiveness, Entrepreneurship and Innovation, under the call RESEARCH - CREATE - INNOVATE (project code: T2EDK-02358).
	
	\bibliographystyle{splncs04}
	\bibliography{references}

\end{document}